\documentclass[10pt, a4paper]{article}
\usepackage{lrec}
\usepackage{multibib}
\newcites{languageresource}{Language Resources}
\usepackage{graphicx}
\usepackage{tabularx}
\usepackage{soul}
\usepackage[colorinlistoftodos]{todonotes}
\usepackage{epstopdf}
\usepackage[latin1]{inputenc}

\usepackage{tipa}

\usepackage{hyperref}
\usepackage{xstring}

\usepackage{gb4e}

\usepackage{framed}

\usepackage{multirow}

\usepackage{booktabs}
\usepackage{amsmath}
\usepackage{multicol}

\title{MCScript: A Novel Dataset for Assessing Machine Comprehension Using Script Knowledge}

\name{Simon Ostermann, Ashutosh Modi, Michael Roth, Stefan Thater, Manfred Pinkal}

\address{Saarland University \\
         Saarbrücken, Germany \\
         \{simono\textpipe ashutosh\textpipe mroth\textpipe stth\textpipe pinkal\}@coli.uni-saarland.de\\}

\abstract{We introduce a large dataset of narrative texts and questions about these texts, intended to be used in a machine comprehension task that requires 
reasoning using commonsense knowledge. Our dataset complements similar datasets in that we focus on stories about everyday activities, such as going to the movies or working in the garden, and that the questions require commonsense knowledge, or more specifically, \emph{script knowledge}, to be answered. We show that our mode of data collection via crowdsourcing results in a substantial amount of such inference questions. The dataset forms the basis of a shared task on commonsense and script knowledge organized at SemEval 2018 and provides challenging test cases for the broader natural language understanding community. \\ \newline \Keywords{machine comprehension, reading comprehension, commonsense knowledge, script knowledge} }

\begin{document}

\setstcolor{red}

\maketitleabstract

\section{Introduction}

Ambiguity and implicitness are inherent properties of natural language that cause challenges for computational models of language understanding. 
% People implicitly 
In everyday communication, people assume a shared common ground which forms a basis for efficiently resolving ambiguities and for inferring implicit information. %based on which ambiguities can efficiently be resolved and implicit information can be inferred.
Thus, recoverable information is often left unmentioned or underspecified. Such information may include encyclopedic and commonsense knowledge. This work focuses on commonsense knowledge about everyday activities, so-called \textit{scripts}.

This paper introduces a dataset to evaluate natural language understanding approaches with a focus on interpretation processes requiring inference based on commonsense knowledge. In particular, we present \textit{MCScript}, a dataset for assessing the contribution of script knowledge to machine comprehension.
%In this paper, we introduce \textit{MCScript:} a dataset to evaluate \textit{Machine Comprehension} systems, with a focus on  interpretation processes requiring \textit{Script} knowledge.
%guides the expectations of the human speaker/listener and also helps to resolve the inherent ambiguities in the language. 
%This paper introduces a dataset for assessing the contribution of common sense knowledge about daily activities in language comprehension. In particular, 
Scripts are sequences of events describing stereotypical human activities (also called scenarios), for example baking a cake or taking a bus \cite{Schank1975}. 
To illustrate the importance of script knowledge, consider Example~(\ref{intro:ex1}):

%Script knowledge plays an important role in linguistic formulation and cognitive processing of natural language (for example, see Chapter 3, \newcite{modiThesis17}). 

\begin{exe}
	\ex The waitress brought Rachel's order. She ate the food with great pleasure. \label{intro:ex1}
\end{exe}

Without using commonsense knowledge, it may be difficult to tell who ate the food: Rachel or the waitress. In contrast, if we utilize commonsense knowledge, in particular, script knowledge about the \textsc{eating in a restaurant} scenario, we can make the following inferences: Rachel is most likely a customer, since she received an order. It is usually the customer, and not the waitress, who eats the ordered food. So \textit{She} most likely refers to Rachel.

Various approaches for script knowledge extraction and processing have been proposed in recent years. However, systems have been evaluated for specific aspects of script knowledge only, such as event ordering \cite{modi:CONLL2014,modi:ICLR2014}, event paraphrasing \cite{regneri2010learning,Wanzare2017} or event prediction (namely, the narrative cloze task \cite{chambers2008unsupervised,chambers2009unsupervised,pichotta2014statistical,pichotta:acl16,modi2016}). These evaluation methods lack a clear connection to real-world tasks. %Recently, there have been efforts to assess the contribution of script knowledge in resolving discourse referents in the text \cite{modiTacl2017}. 
Our MCScript dataset provides an extrinsic evaluation framework, based on text comprehension involving commonsense knowledge. This framework makes it possible to assess system performance in a multiple-choice question answering setting, without imposing any specific structural or methodical requirements. %To bridge this gap, we propose a novel data set that directly relates commonsense knowledge and natural language comprehension. We present a data set for machine comprehension using commonsense knowledge. 

\begin{figure}
	\begin{framed}
		\begin{tabularx}{\columnwidth}{p{.2cm}XX}
			\textbf{T} & \multicolumn{2}{p{6.5cm}}{I wanted to plant a tree. I went to the home and garden store and picked a nice oak. Afterwards, I planted it in my garden.}\\
			& & \\
			\textbf{Q1} & \multicolumn{2}{l}{What was used to dig the hole?}\\
			& a. a shovel & b. his bare hands\\
			& & \\
			\textbf{Q2} & \multicolumn{2}{l}{When did he plant the tree?}\\
			& a. after watering it & b. after taking it home\\
		\end{tabularx}
	\end{framed}
	\caption{An example for a text snippet with two reading comprehension questions.}
	\label{ex1}
\end{figure}

MCScript is a collection of (1) narrative texts, (2) questions of various types referring to these texts, and (3) pairs of answer candidates for each question. It comprises approx.~2,100 texts and a total of approx.~14,000 questions. Answering a substantial subset of questions requires knowledge beyond the facts mentioned in the text, i.e. it requires inference using commonsense knowledge about everyday activities. An example is given in Figure \ref{ex1}. For both questions, the correct choice for an answer requires commonsense knowledge about the activity of planting a tree, which goes beyond what is mentioned in the text. Texts, questions, and answers were obtained through crowdsourcing. In order to ensure high quality, we manually validated and filtered the dataset. Due to our design of the data acquisition process, we ended up with a substantial subset of questions that require commonsense inference (27.4\%). %Answering some of the questions requires knowledge beyond the facts mentioned in the text. In particular, a substantial subset of questions requires inference using commonsense knowledge via scripts. 

\section{Corpus}
Machine comprehension datasets consist of three main components: texts, questions and answers. In this section, we describe our data collection for these 3 components. We first describe a series of pilot studies that we conducted in order to collect commonsense inference questions (Section \ref{subsec:pilot}). In Section~\ref{subsec:datacollection}, we discuss the resulting data collection of questions, texts and answers via crowdsourcing on Amazon Mechanical Turk\footnote{\url{www.mturk.com}} (henceforth \textit{MTurk}). Section \ref{subsec:selection} gives information about some necessary postprocessing steps and the dataset validation. Lastly, Section \ref{subsec:statistics} gives statistics about the final dataset. 

\subsection{Pilot Study}
\label{subsec:pilot}
As a starting point for our pilots, we made use of texts from the \textit{InScript} corpus \cite{modiinscript}, which provides stories centered around everyday situations (see Section~\ref{subsubsec:texts}). %Since InScript contains texts for 10 scenarios only, we later collected a corpus that contains 100 additional different scenarios, following the collection approach of InScript, which is further described in Sections \ref{subsubsec:scenarios} (scenario selection) and \ref{subsubsec:texts} (text collection). 
We conducted three different pilot studies to determine the best way of collecting questions that require inference over commonsense knowledge:

The most intuitive way of collecting reading comprehension questions is to show texts to workers and let them formulate questions and answers on the texts, which is what we tried internally in a \textit{first pilot}. Since our focus is to provide an evaluation framework for inference over commonsense knowledge, we manually assessed the number of questions that indeed require common sense knowledge. We found too many questions and answers collected in this manner to be lexically close to the text. 

In a \textit{second pilot}, we investigated the option to take the questions collected for one text and show them as questions for another text of the same scenario. While this method resulted in a larger number of questions that required inference, we found the majority of questions to not make sense at all when paired with another text. Many questions were specific to a text (and not to a scenario), requiring details that could not be answered from other texts.% 

Since the two previous pilot setups resulted in questions that centered around the texts themselves, we decided for a \textit{third pilot} to not show workers any specific texts at all. Instead, we asked for questions that centered around a specific script scenario (e.g. \textsc{eating in a restaurant}). We found this mode of collection to result in questions that have the right level of specificity for our purposes: namely, questions that are related to a scenario and that can be answered from different texts (about that scenario), but for which a text does not need to provide the answer explicitly.

The next section will describe the mode of collection chosen for the final dataset, based on the third pilot, in more detail.

\subsection{Data Collection}	
\label{subsec:datacollection}

\subsubsection{Scenario Selection}
\label{subsubsec:scenarios}
As mentioned in the previous section, we decided to base the question collection on script scenarios rather than specific texts. As a starting point for our data collection, we use scenarios from three script data collections \cite{regneri2010learning,Singh2002,Wanzare2016}. Together, these resources contain more than 200 scenarios. To make sure that scenarios have different complexity and content, we selected 80 of them and came up with 20 new scenarios. Together with the 10 scenarios from InScript, we end up with a total of 110 scenarios.
%In the next step, we grouped scenarios according to their general theme, such as \textit{chores} (e.g. \textsc{doing laundry} or \textsc{washing dishes}), or \textit{leisure time activities} (e.g. \textsc{making a bonfire} or \textsc{taking a bath}). 

%We selected a mixture of scenarios from each group, ending up with 39 scenarios from \textit{DeScript} (10 of which are also used in the \textit{InScript} corpus), 50 scenarios from \textit{OMCS}, and one scenario from \textit{RKP}. Additionally, we added 20 completely new scenarios. %We also created 10 new scenarios that would bridge gaps between other scenarios: For \textsc{planting a tree} and \textsc{mowing the lawn}, we for example introduced a new scenario \textsc{working in the garden}, connecting the former two. Another 10 were added by formulating scenarios for a number of randomly selected ROC stories \cite{mostafazadeh2016corpus}.

%The collection resulted in 110 scenarios 

\subsubsection{Texts}
\label{subsubsec:texts}
For the collection of texts, we followed \newcite{modiinscript}, where workers were asked to write a story about a given activity ``as if explaining it to a child''. This results in elaborate and explicit texts that are centered around a single scenario. Consequently, the texts are syntactically simple, facilitating machine comprehension models to focus on semantic challenges and inference. We collected 20 texts for each scenario. Each participant was allowed to write only one story per scenario, but work on as many scenarios as they liked. For each of the 10 scenarios from InScript, we randomly selected 20 existing texts from that resource.%For the 10 scenarios also used in %\todo[inline]{Not sure if you should italicize InScript again and again}
%\textit{InScript}, we just selected  rather than collecting more.

%TODO: Some MTurk statistics?

\subsubsection{Questions}
\label{subsubsec:questions}
For collecting questions, workers were instructed to ``imagine they told a story about a certain scenario to a child and want to test if the child understood everything correctly''. This instruction also ensured that questions are linguistically simple, elaborate and explicit. Workers were asked to formulate questions about details of such a situation, i.e.~independent of a concrete narrative. This resulted in questions, the answer to which is not literally mentioned in the text.  %While collecting questions directly based on a text might seem more intuitive, we found this to primarily result in questions that are lexically close to the text. Collecting questions that way might also not guarantee that commonsense inference was required. %, which is a major difference to other available machine comprehension data sets such as \textit{MCTest} TODO: Not anymore! TriviaQA do the same!

To cover a broad range of question types, we asked participants to write 3 temporal questions (asking about time points and event order), 3 content questions (asking about persons or details in the scenario) and 3 reasoning questions (asking how or why something happened). They were also asked to formulate 6 free questions, which resulted in a total of 15 questions. %\todo[inline]{May be rephrase the next line or may be even remove it as the subsequent line says what you want to say}While this number might sound high, we found that it is necessary in order to avoid that participants only write down obvious questions for a scenario. 
Asking each worker for a high number of questions enforced that more creative questions were formulated, which go beyond obvious questions for a scenario. %, participants were encouraged to be creative.% and write less obvious questions. 

Since participants were not shown a concrete story, we asked them to use the neutral pronoun ``they'' to address the protagonist of the story. %Participants were paid \$0.50 per 15 questions. %\todo[inline]{May be rephrase the next line a bit, it looks a bit convoluted}
We permitted participants to work on as many scenarios as desired and we collected questions from 10 participants per scenario.
%When collecting questions independent of a concrete narrative text, they have to manually be mapped to matching texts, i.e.  to texts that provide sufficient information for finding or inferring an answer. We use trained expert annotators for identifying all questions that are relevant to a given text. As in the pilot study (see Section~\ref{sec:pilot}), they are going to mark for question-text pairs whether a question is \textit{answerable}, \textit{not answerable} or \textit{inferable} by commonsense and the text. 

\subsubsection{Answers}
\label{subsubsec:answers}
Our mode of question collection results in questions that are not associated with specific texts. %In order to crowdsource answers, we thus conducted a series of experiments, in which each text was first paired with 15 randomly selected questions from the same scenario. 
For each text, we collected answers for 15 questions that were randomly selected from the same scenario. Since questions and texts were collected independently, answering a random question is not always possible for a given text. %, or is only possible by inferring facts not mentioned in the text. To make these differences apparent, 
Therefore, we carried out answer collection in two steps. In the first step, we asked participants to assign a category to each text--question pair.
%\begin{enumerate}
%	\item The question can be answered from the text.
%	\item The question can be answered using common-sense knowledge.
%	\item The question cannot be answered, even through inference by using commonsense knowledge.
%	\item The question does not make sense at all with the given text.
%\end{enumerate}

We distinguish two categories of answerable questions: The category \textit{text-based} was assigned to questions that can be answered from the text directly. If the answer could only be inferred by using commonsense knowledge, the category \textit{script-based} was assigned. Making this distinction is interesting for evaluation purposes, since it enables us to estimate the number of commonsense inference questions. For questions that did not make sense at all given a text, \textit{unfitting} was assigned. If a question made sense for a text, but it was impossible to find an answer, the label \textit{unknown} was used.
%In case a question cannot be answered given a text, \textit{unknown} was assigned if the worker could not find an appropriate answer, and \textit{unfitting} was used for questions that did not make sense at all given the text.

In a second step, we told participants to formulate a plausible correct and a plausible incorrect answer candidate to answerable questions (\textit{text-based} or \textit{script-based}). To level out the effort between answerable and non-answerable questions, participants had to write a new question when selecting \textit{unknown} or \textit{unfitting}.%, which was however not further used in the scope of this corpus creation.

In order to get reliable judgments about whether or not a question can be answered, we collected data from 5 participants for each question and decided on the final category via majority vote (at least 3 out of 5). Consequently, for each question with a majority vote on either \textit{text-based} or \textit{script-based}, there are 3 to 5 correct and incorrect answer candidates, one from each participant who agreed on the category. Questions without a clear majority vote or with ties were not included in the dataset.

%For each text with 15 random questions, we collected data from 5 different participants, paying \$0.60 to each participant.

\begin{table}
	\begin{tabular}{|cc||cc|}
		\hline
		\multicolumn{2}{|c||}{\textbf{answerable}} & \multicolumn{2}{c|}{\textbf{not answerable}}\\
		\hline
		\textbf{text-based}& \textbf{script-based} & \textbf{unknown} & \textbf{unfitting}\\
		10,160 & 3,914 & 9,974 & 3,172\\
		\hline
		\multicolumn{2}{|c||}{14,074} & \multicolumn{2}{c|}{13,246}\\
		\hline
	\end{tabular}
	\caption{Distribution of question categories}
	\label{tab}
\end{table}

\subsubsection{Data Post-Processing}
\label{subsubsec:postprocessing}
We performed four post-processing steps on the collected data. 

\begin{itemize}
	\item We manually filtered out texts that were instructional rather than narrative.
	\item All texts, questions and answers were spellchecked by running aSpell\footnote{\url{http://aspell.net/}} and manually inspecting all corrections proposed by the spellchecker.
	\item We found that some participants did not use ``they'' when referring to the protagonist. We identified ``I'', ``you'', ``he'', ``she'', ``my'', ``your'', ``his'', ``her'' and ``the person'' as most common alternatives and replaced each appearance manually with ``they'' or ``their'', if appropriate.
	\item We manually filtered out invalid questions, e.g. questions that are suggestive (``Should you ask an adult before using a knife?'') or that ask for the personal opinion of the reader (``Do you think going to the museum was a good idea?'').
\end{itemize}

%In a second step, we manually examined and cleaned up the data set. 

%Finally, we merged obvious paraphrases among the correct and among incorrect candidates respectively. For this, we normalized every answer candidate by  After transformation, all string identical answers were merged. Answers that were contained literally in another answer were merged into the larger one. Finally, all answers with a Levenshtein distance \cite{levenshtein1966binary} of 3 or less that did not contain "yes", "no", or numbers were also merged.
%In order to select a single correct and a single incorrect answer out of the merged candidates, we then tried to identify answers in a way that the data set is more challenging for simple (overlap) systems: We selected the correct answer candidate with the lowest overlap to the text, and the incorrect answer candidate with the highest overlap to the text.

%While this selection criterion resulted in good correct answers, we found after a manual inspection of 100 randomly picked questions that many inappropriate incorrect answers were selected, e.g. because they were actually good answers: The participant who wrote it made a mistake.

%To reduce noise in the data, we thus decided  to also rely on the majority opinion for incorrect answers: We take the incorrect answer candidate that was merged most often with another one. If all answers are unique, i.e. none of them was merged, we fall back to the overlap criterion, selecting the answer candidate with the highest overlap to the text.

\subsection{Answer Selection and Validation}
\label{subsec:selection}
We finalized the dataset by selecting one correct and one incorrect answer for each question--text pair. To increase the proportion of non-trivial inference cases, we chose the candidate with the \textit{lowest} lexical overlap with the text from the set of correct answer candidates as \textit{correct} answer. Using this principle also for incorrect answers leads to problems. We found that many incorrect candidates were not plausible answers to a given question. Instead of selecting a candidate based on overlap, we hence decided to rely on
%When choosing the candidate with the highest overlap to the text, we observed many cases in which an improper answer was selected, because many incorrect answers were nonsensical. 
%We thus relied 
majority vote and selected the candidate from the set of incorrect answers that was most often mentioned. 

For this step, we normalized each candidate by lowercasing, deleting punctuation and stop words (articles, \textsl{and}, \textsl{to} and \textsl{or}), and transforming all number words into digits, using \textit{text2num}\footnote{\url{https://github.com/ghewgill/text2num}}. We merged all answers that were string-identical, contained another answer, or had a Levenshtein distance \cite{levenshtein1966binary} of 3 or less to another answer. The ``most frequent answer'' was then selected based on how many other answers it was merged with.
%Note that this step was only necessary to find trivial paraphrases for the majority-based selection of a candidate (cf. next section). After the selection of a candidate, the normalizations were undone.
% (based on the normalizing outlined in \ref{subsubsec:postprocessing}, ). 
Only if there was no majority, we selected the candidate with the highest overlap with the text as a fallback. %These selection criteria ensure that (a) the correct answer is challenging to identify using simple word overlap metrics and (b) the incorrect answer is a plausible answer, as indicated by the fact that multiple turkers mentioned it.

% While this selection criterion results in appropriate correct answers, we found after a manual inspection of 100 randomly sampled questions that many inappropriate incorrect answers were selected, e.g. because they were actually good answers: The participant who wrote it made a mistake. We thus decided to rely on the majority opinion for incorrect answers, by choosing the incorrect candidate that a majority of participants had mentioned\footnote{We identified trivial paraphrases as described above. Thus, the majority candidate is the one that was most often merged with another candidate.}. Only if there was no clear majority, we chose the candidate with the highest overlap to the text as a fallback.
%	After another random sampling of 100 questions, we found that an approximate number of 5-10\% of answers was still inappropriate. 
Due to annotation mistakes, we found a small number of chosen correct and incorrect answers to be inappropriate, that is, some ``correct'' answers were actually incorrect and vice versa. Therefore, we manually validated the complete dataset in a final step. We asked annotators to read all texts, questions, and answers, and to mark for each question whether the correct and incorrect answers were appropriate. If an answer was inappropriate or contained any errors, they selected a different answer from the set of collected candidates. For approximately 11.5\% of the questions, at least one answer was replaced. 135 questions (approx.\ 1\%) were excluded from the dataset because no appropriate correct or incorrect answer could be found.

\subsection{Data Statistics}
\label{subsec:statistics}
%\begin{figure}
%\centering
%\vspace{-0.5cm}
%\includegraphics[width=.8\columnwidth]{figures/question-categories.pdf}
%\vspace{-1.5cm}
%	\caption{Distribution of question categories, based on majority voting from 5 participants.}
%	\label{fig:question-categories}
%\end{figure}

For all experiments, we admitted only experienced MTurk workers who are based in the US. One HIT\footnote{A \textit{Human Intelligence Task} (HIT) is one single experimental item in MTurk.} consisted of writing one text for the text collection, formulating 15 questions for the question collection, or finding 15 pairs of answers for the answer collection. We paid \$0.50 per HIT for the text and question collection, and \$0.60 per HIT for the answer collection.

More than 2,100 texts were paired with 15 questions each, resulting in a total number of approx.\ 32,000 annotated questions. For 13\% of the questions, the workers did not agree on one of the 4 categories with a 3 out of 5 majority, so we did not include these questions in our dataset. 

The distribution of category labels on the remaining 87\% is shown in Table \ref{tab}. 14,074 (52\%) questions could be answered. Out of the answerable questions, 10,160 could be answered from the text directly (\textit{text-based}) and 3,914 questions required the use of commonsense knowledge (\textit{script-based}). After removing 135 questions during the validation, the final dataset comprises 13,939 questions,~3,827 of which require commonsense knowledge (i.e.~27.4\%). This ratio was manually verified based on a random sample of questions.

We split the dataset into training (9,731 questions on 1,470 texts), development (1,411 questions on 219 texts), and test set (2,797 questions on 430 texts). Each text appears only in one of the three sets. The complete set of texts for 5 scenarios was held out for the test set. %All texts for 5 scenarios were withdrawn exclusively for the test set, the rest was selected randomly.

The average text, question, and answer length is 196.0 words, 7.8 words, and 3.6 words, respectively. On average, there are 6.7 questions per text. %Each scenario is represented with 15 texts at minimum.

%Figure \ref{fig:question-categories} shows the distribution of question categories on this number, as determined by the majority vote of 5 MTurk participants, on all annotated questions. As can be seen, 46\% of all presented questions could be answered, i.e. form the final data set. For 13\%, there was no majority for one category. Among all answerable questions, roughly a third are questions that require commonsense knowledge for answering.%\footnote{We are planning to verify this number with an expert annotation.} This high number indicates that collecting questions and texts independently indeed results in a high amount of inference questions.

\begin{figure}
	\centering
	\vspace{-0.5cm}
	\includegraphics[width=\columnwidth]{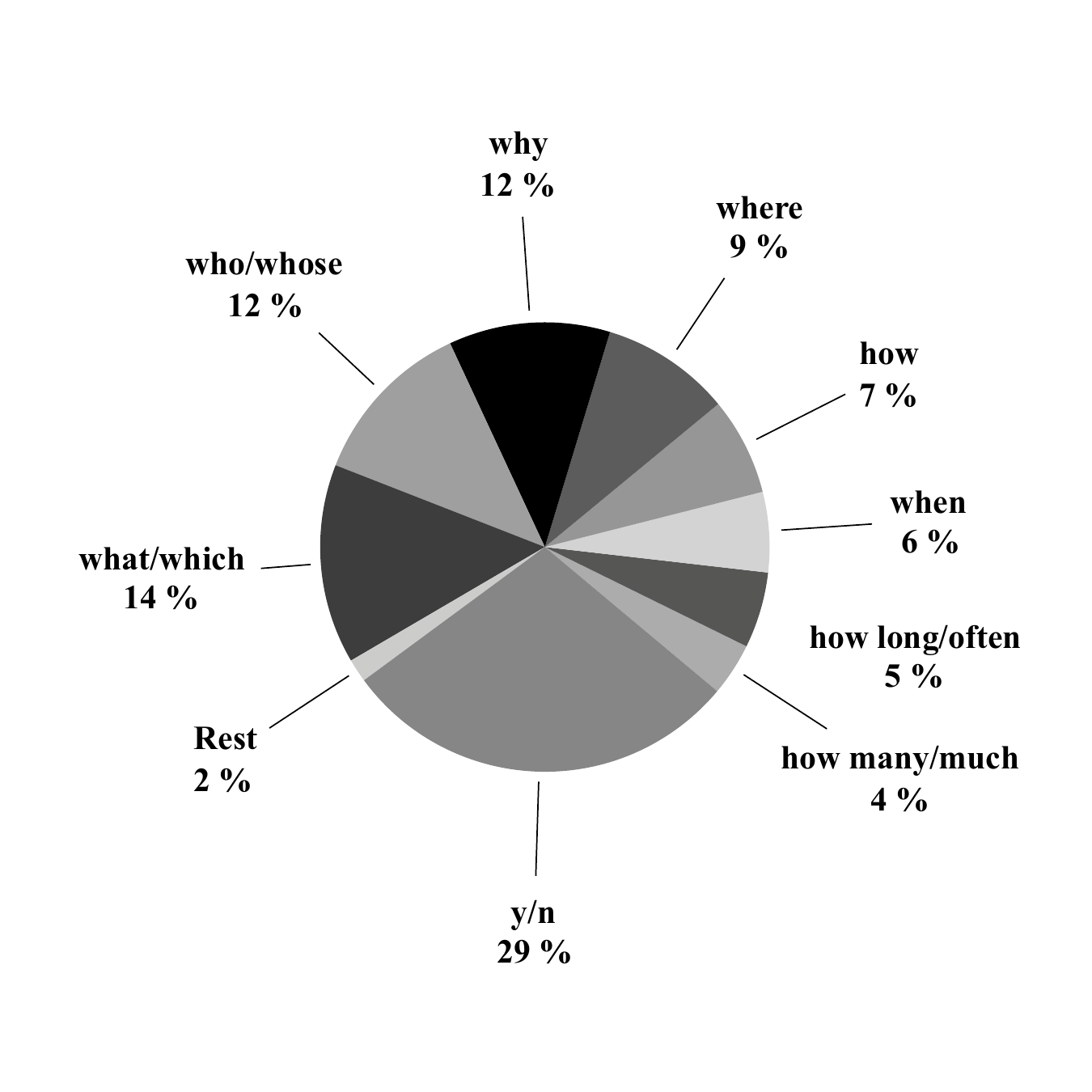}
	\vspace{-.5cm}
	\caption{Distribution of question types in the data.}
	\label{fig:question-types}
\end{figure}
%After cleaning the data by excluding flawed or inappropriate texts and questions (as described above), the final corpus contains approx. 2100 texts with a total of 14,000 questions, i.e. there are about 6.7 questions per text. 

Figure \ref{fig:question-types} shows the distribution of question types in the dataset, which we identified using simple heuristics based on the first words of a question: Yes/no questions were identified as questions starting with an auxiliary or modal verb, all other question types were determined based on the question word. 

%\textbf{[TODO: Hiwi annotations (validation of cs question rate)], more details. Number of scenarios, texts per scenario, text length, question length, answer length, etc}

We found that 29\% of all questions are yes/no questions. Questions about details of a situation (such as \textit{what}/ \textit{which} and \textit{who}) form the second most frequent question category. %Participant and coreference information extracted from script knowledge will help in answering these questions. 
Temporal questions (\textit{when} and \textit{how long/often}) form approx.~11\% of all questions. %Since temporal ordering information is explicitly encoded in script knowledge, we expect that it will especially help in answering these questions. Examples for some questions can be found in the next Section.
We leave a more detailed analysis of question types for future work.

\section{Data Analysis}
%One of the main aims of our effort was to come up with inference questions that do not repeat passages in the text. 

As can be seen from the data statistics, our mode of collection leads to a substantial proportion of questions that require inference using commonsense knowledge. Still, the dataset contains a large number of questions in which the answer is explicitly contained or implied by the text: Figure \ref{ex:text} shows passages from an example text of the dataset together with two such questions. For question Q1, the answer is given literally in the text. Answering question Q2 is not as simple; it can be solved, however, via standard semantic relatedness information (chicken and hotdogs are meat; water, soda and juice are drinks).

\begin{figure}
	\begin{framed}
		\begin{tabularx}{\columnwidth}{p{.2cm}p{.1cm}Xp{.1cm}X}
			\textbf{T} & \multicolumn{4}{p{6.5cm}}{It was time to prepare for the picnic that we had plans for the last couple weeks.  \dots  %First, I gathered all of the things that I would need for the picnic table. %That included a plastic cover for the table, paper plates, plastic cups, plastic utensils, napkins, aluminum foil and plastic bags that would be used for garbage and taking things back home. Next 
				I needed to set up the cooler, which included bottles of water, soda and juice to keep everyone hydrated. Then I needed to ensure that we had all the food we intended to bring or cook. So at home, I prepared baked beans, green beans and macaroni and cheese. \dots %These foods were put into separate aluminum containers, and covered with aluminum fall so they could be transported to the picnic area and ready to be served. 
				But in a cooler, I packed chicken, hotdogs, hamburgers and rots that were to be cooked on the grill once we were at the picnic location.}\\
			& & & & \\
			\textbf{Q1} & \multicolumn{4}{l}{What did they bring to drink?}\\
			& a. &Water, soda and juice. & b. &Water, wine coo\-lers and sports drinks.\\
			& & & &\\
			\textbf{Q2} & \multicolumn{4}{l}{What type of food did they pack?}\\
			& a. &Meat, drinks and side dishes. & b. &Pasta salad only.\
		\end{tabularx}
	\end{framed}
	\caption{An example text with 2 questions from MCScript}
	\label{ex:text}
\end{figure}

%\begin{exe}
%	\ex It was time to prepare for the picnic that we had plans for the last couple weeks.  \dots  %First, I gathered all of the things that I would need for the picnic table. %That included a plastic cover for the table, paper plates, plastic cups, plastic utensils, napkins, aluminum foil and plastic bags that would be used for garbage and taking things back home. Next 
%	I needed to set up the cooler, which included bottles of water, soda and juice to keep everyone hydrated. Then I needed to ensure that we had all the food we intended to bring or cook. So at home, I prepared baked beans, green beans and macaroni and cheese. \dots %These foods were put into separate aluminum containers, and covered with aluminum fall so they could be transported to the picnic area and ready to be served. 
%	But in a cooler, I packed chicken, hotdogs, hamburgers and rots that were to be cooked on the grill once we were at the picnic location.\\
%		\textbf{Q1}: What did they bring to drink?\\
%	a. Water, soda and juice.\\
%	b. Water, wine coolers and sports drinks.\\
%	\textbf{Q2}: What type of food did they pack?\\
%	a. Meat, drinks and side dishes.\\
%	b. Pasta salad only.
%	\label{ex:text}
%\end{exe}

The following cases require commonsense inference to be decided. In all these cases, the answers are not overtly contained nor easily derivable from the respective texts. We do not show the full texts, but only the scenario names for each question.

\begin{exe}
	\ex \textsc{borrowing a book from the library}\\
	Did they have to pay anything to borrow the book?\\
	a. yes\\
	b. no
	\label{ex:library}
\end{exe}

\begin{exe}
	\ex \textsc{changing a baby diaper}\\
	Did they throw away the old diaper?\\
	a. Yes, they put it into the bin.\\
	b. No, they kept it for a while.
	\label{ex:diaper}
\end{exe}

\begin{exe}
	\ex \textsc{cleaning the table}\\
	When did they clean the table?\\
	a. After a meal\\
	b. Before they ate
	\label{ex:temporal}
\end{exe}

\begin{exe}
	\ex \textsc{preparing a picnic}\\
	Who is packing the picnic?\\
	a. the children\\
	b. the parents
	\label{ex:who}
\end{exe}

\begin{exe}
	\ex \textsc{taking a shower} \\
	How long did the shower take?\\
	a. a few hours\\
	b. a few minutes
	\label{ex:shower}
\end{exe}

Example \ref{ex:library} refers to a library setting. Script knowledge helps in assessing that usually, \textit{paying} is not an event when borrowing a book, which answers the question. Similarly, event information helps in answering the questions in Examples \ref{ex:diaper} and \ref{ex:temporal}. %In Example \ref{ex:diaper}, script knowledge should contain the information that \textit{throwing old diapers away} is an event within the \textsc{changing diapers} script. 
In Example \ref{ex:who}, knowledge about the typical role of parents in the preparation of a picnic will enable a plausibility decision. %In Example \ref{ex:temporal}, script knowledge helps in assessing that after eating, a table is usually cleaned.
Similarly, in Example \ref{ex:shower}, it is commonsense knowledge that showers usually take a few minutes rather than hours. 

\begin{exe}
	\ex \textsc{making breakfast} \\
	What time of the day is breakfast eaten?\\
	a. at night\\
	b. in the morning
	\label{ex:breakfast}
\end{exe}

There are also cases in which the answer can be inferred from the text, but where commonsense knowledge is still beneficial: The text for example \ref{ex:breakfast} does not contain the information that breakfast is eaten in the morning, but it could still be inferred from many pointers in the text (e.g.~phrases such as \textit{I woke up}), or from commonsense knowledge.

These few examples illustrate that our dataset covers questions with a wide spectrum of difficulty, from rather simple questions that can be answered from the text to challenging inference problems.

\section{Experiments}
In this section, we assess the performance of baseline models on MCScript, using accuracy as the evaluation measure. We employ models of differing complexity: two unsupervised models using only word information and distributional information, respectively, and two supervised neural models. We assess performance on two dimensions: One, we show how well the models perform on text-based questions as compared to questions that require common sense for finding the correct answer. Two, we evaluate each model for each different question type.

\subsection{Models}
\subsubsection*{Word Matching Baseline}
We first use a simple word matching baseline, by selecting the answer that has the highest literal overlap with the text. In case of a tie, we randomly select one of the answers. 
\subsubsection*{Sliding Window}
The second baseline is a sliding window approach that looks at windows of $w$ tokens on the text. Each text and each answer are represented as a sequence of word embeddings. The embeddings for each window of size $w$ and each answer are then averaged to derive window and answer representations, respectively. The answer with the lowest cosine distance to one of the windows of the text is then selected as correct.
\subsubsection*{Bilinear Model}
We employ a simple neural model as a third baseline. In this model, each text, question, and answer is represented by a vector. For a given sequence of words  $w_1 \ldots w_n$, we compute this representation by averaging over the components of the word embeddings $\textbf{w}_i$ that correspond to a word $w_i$, and then apply a linear transformation using a weight matrix. This procedure is applied to each answer $a$ to derive an answer representation $\textbf{a}$. The representation of a text $\textbf{t}$ and of a question $\textbf{q}$ are computed in the same way. We use different weight matrices for $\textbf{a}$, $\textbf{t}$ and $\textbf{q}$, respectively. A combined representation $\textbf{p}$ for the text--question pair is then constructed using a bilinear transformation matrix $\textbf{W}$:
\begin{equation}
\textbf{p} = \textbf{t}^{\top}\textbf{W}\textbf{q}
\end{equation}

We compute a score for each answer by using the dot product and pass the scores for both answers through a softmax layer for prediction. The probability $p$ for an answer $a$ to be correct is thus defined as:

\begin{equation}
p(a|t,q) = softmax(\textbf{p}^{\top} \textbf{a})
\end{equation}

\subsubsection*{Attentive Reader}
The attentive reader is a well-established machine comprehension model that reaches good performance e.g.~on the \textit{CNN/Daily Mail} corpus \cite{hermann2015teaching,chen2016thorough}. We use the model formulation by \newcite{chen2016thorough} and \newcite{lai2017race}, who employ bilinear weight functions to compute both attention and answer-text fit. Bi-directional GRUs are used to encode questions, texts and answers into hidden representations. For a question $q$ and an answer $a$, the last state of the GRUs, $\textbf{q}$ and $\textbf{a}$, are used as representations, while the text is encoded as a sequence of hidden states $\textbf{t}_1...\textbf{t}_n$. We then compute an attention score $s_j$ for each hidden state $\textbf{t}_j$ using the question representation $\textbf{q}$, a weight matrix $\textbf{W}_a$, and an attention bias $b$. Last, a text representation $\textbf{t}$ is computed as a weighted average of the hidden representations:
\begin{equation}
\begin{split}
s_j =& softmax_j(\textbf{t}_j^{\top} \textbf{W}_a \textbf{q} + b) \\
\textbf{t} =& \sum_j s_j \textbf{t}_j
\end{split}
\end{equation}
The probability $p$ of answer $a$ being correct is then predicted using another bilinear weight matrix $\textbf{W}_s$, followed by an application of the softmax function over both answer options for the question:
\begin{equation}
p(a|t,q) = softmax(\textbf{t}^{\top}\textbf{W}_s\textbf{a})
\end{equation}
%\subsubsection*{Parallel-Hierarchical Model}
%\newcite{trischlerparallel} propose a new network architecture for machine comprehension on small datasets, the parallel-hierarchical model. The core idea is to represent text, questions and answers on various linguistic levels: On a semantic level, a sentential level, and word-by-word using sliding windows. Also, they employ empirically driven estimations of weight initializations that let the training converge on small data sets. They achieve the current state of the art on MCTest \cite{richardson2013mctest}, a frequently used machine comprehension benchmark dataset.

\subsection{Implementation Details}
Texts, questions and answers were tokenized using NLTK\footnote{\url{http://www.nltk.org/}} and lowercased. We used 100-dimensional GloVe vectors\footnote{\url{https://nlp.stanford.edu/projects/glove/}} \cite{pennington2014glove} to embed each token. For the neural models, the embeddings are used to initialize the token representations, and are refined during training. For the sliding similarity window approach, we set $w=10$. 

The vocabulary of the neural models was extracted from training and development data.  For optimizing the bilinear model and the attentive reader, we used vanilla stochastic gradient descent with gradient clipping, if the norm of gradients exceeds 10. The size of the hidden layers was tuned to $64$, with a learning rate of $0.2$, for both models. We apply a dropout of $0.5$ to the word embeddings. Batch size was set to $25$ and all models were trained for 150 epochs. During training, we measured performance on the development set, and we selected the model from the
best performing epoch for testing.%\marginpar{order?)}
%\textbf{Methods: bilinear classifier, attentive reader, some state-of-the-art model? HUB!}
\begin{table}[t]
	\centering
	\begin{tabular}{lrrr}
		\toprule
		\textit{Model}&\textit{Text}&\textit{CS}&\textit{Total}\\
		\midrule
		\midrule
		Chance & 50.0& 50.0& 50.0\\
		\midrule
		Word Overlap & 41.8& 59.0& 54.4\\
		Sliding Window & 55.7& 53.1& 55.0\\
		Bilinear Model & 69.8 & 71.4 & 70.2\\
		Attentive Reader & \textbf{70.9} & \textbf{75.2}& \textbf{72.0}\\
		%Parallel-Hierarchical & & X& X\\
		%\midrule
		%Human upper bound & 98.2& X& X\\
		\midrule	
		Human Performance & & & 98.2 \\
		\bottomrule
	\end{tabular}
	\caption{Accuracy of the baseline systems on text-based (\textit{Text}), on commonsense-based questions (\textit{CS}), and on the whole test set (\textit{Total}). All numbers are percentages.}
	\label{tab:results}
\end{table}

\subsection{Results and Evaluation}
\subsubsection*{Human Upper Bound}
As an upper bound for model performance, we assess how well humans can solve our task. Two trained annotators labeled the correct answer on all instances of the test set. They agreed with the gold standard in 98.2\ \% of cases. This result shows that humans have no difficulty in finding the correct answer, irrespective of the question type.

\begin{figure*}[t]
	\centering
	\includegraphics[width=.8\textwidth]{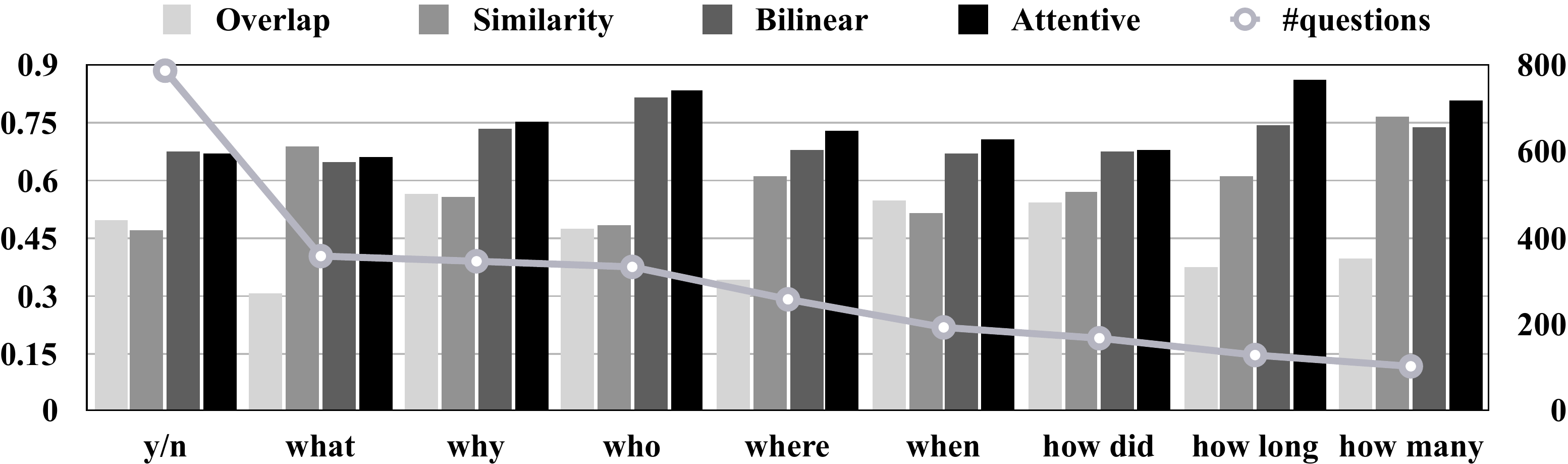}
	\caption{Accuracy values of the baseline models on question types appearing $>25$ times.}
	\label{fig:eval}
\end{figure*}

\subsubsection*{Performance of the Baseline Models}
Table \ref{tab:results} shows the performance of the baseline models as compared to the human upper bound and a random baseline. As can be seen, neural models have a clear advantage over the pure word overlap baseline, which performs worst, with an accuracy of $54.4\%$.

The low accuracy is mostly due to the nature of correct answers in our data: Each correct answer has a low overlap with the text by design. Since the overlap model selects the answer with a high overlap to the text, it does not perform well. In particular, this also explains the very bad result on text-based questions. The sliding similarity window model does not outperform the simple word overlap model by a large margin: Distributional information alone is insufficient to handle complex questions in the dataset.

Both neural models outperform the unsupervised baselines by a large margin. When comparing the two models, the attentive reader is able to beat the bilinear model by only $1.8\%$. A possible explanation for this is that the attentive reader only attends to the text. Since many questions cannot be directly answered from the text, the attentive reader is not able to perform significantly better than a simpler neural model.

What is surprising is that the attentive reader works better on commonsense-based questions than on text questions. This can be explained by the fact that many commonsense questions do have prototypical answers within a scenario, irrespective of the text. The attentive reader is apparently able to just memorize these prototypical answers, thus achieving higher accuracy. 

Inspecting attention values of the attentive reader, we found that in most cases, the model is unable to properly attend to the relevant parts of the text, even when the answer is literally given in the text. A possible explanation is that the model is confused by the large amount of questions that cannot be answered from the text directly, which might confound the computation of attention values.

Also, the attentive reader was originally constructed for reconstructing literal text spans as answers. Our mode of answer collection, however, results in many correct answers that cannot be found verbatim in the text. This presents difficulties for the attention mechanism.

The fact that an attention model outperforms a simple bilinear baseline only marginally shows that MCScript poses a new challenge to machine comprehension systems. Models concentrating solely on the text are insufficient to perform well on the data.

\subsubsection*{Performance on Question Types}
Figure \ref{fig:eval} gives accuracy values of all baseline systems on the most frequent question types (appearing \textgreater 25 times in the test data), as determined based on the question words (see Section~\ref{subsec:statistics}). The numbers depicted on the left-hand side of the y-axis represent model accuracy. The right-hand side of the y-axis indicates the number of times a question type appears in the test data.

The neural models unsurprisingly outperform the other models in most cases, and the difference for \textit{who} questions is largest. A large number of these questions ask for the narrator of the story, who is usually not mentioned literally in the text, since most stories are written in the first person.

It is also apparent that all models perform rather badly on \textit{yes/no} questions. Each model basically compares the answer to some representation of the text. For yes/no questions, this makes sense for less than half of all cases. For the majority of yes/no questions, however, answers consist only of \textit{yes} or \textit{no}, without further content words.%This method is suboptimal for yes/no questions because the answer often does not contain many meaningful content words. This result emphasizes that models need to treat yes/no questions differently.

\section{Related Work}
\label{sec:relatedwork}
In recent years, a number of reading comprehension datasets have been proposed, including \textit{MCTest} \cite{richardson2013mctest}, \textit{BAbI} \cite{weston2015towards}, the \textit{Children's Book Test} (\textit{CBT}, \newcite{hill2015goldilocks}), \textit{CNN/Daily Mail} \cite{hermann2015teaching}, the \textit{Stanford Question Answering Dataset} (\textit{SQuAD}, \newcite{rajpurkar2016squad}), and \textit{RACE} \cite{lai2017race}. These datasets differ with respect to text type (Wikipedia texts, examination texts, etc.), mode of answer selection (span-based, multiple choice, etc.)~and test systems regarding different aspects of language understanding, but they do not explicitly address commonsense knowledge.

Two notable exceptions are the \textit{NewsQA} and \textit{TriviaQA} datasets. \textit{NewsQA} \cite{trischler2017newsqa} is a dataset of newswire texts from CNN with questions and answers written by crowdsourcing workers.  \textit{NewsQA} closely resembles our own data collection with respect to the method of data acquisition. As for our data collection, full texts were not shown to workers as a basis for question formulation, but only the text's title and a short summary, to avoid literal repetitions and support the generation of non-trivial questions requiring background knowledge.  The NewsQA text collection differs from ours in domain and genre (newswire texts vs.\ narrative stories about everyday events). Knowledge required to answer the questions is mostly factual knowledge and script knowledge is only marginally relevant.    %Also, news wire texts differ from narratives with respect to writing style and linguistic complexity. 
Also, the task is not exactly question answering, but identification of document passages containing the answer.

\textit{TriviaQA} \cite{JoshiTriviaQA2017} is a corpus that contains automatically collected question-answer pairs from 14 trivia and quiz-league websites, together with web-crawled evidence documents from \textit{Wikipedia} and \textit{Bing}. While a majority of questions require world knowledge for finding the correct answer, it is mostly factual knowledge.
%\textbf{
%[TODO: more details, more structure?]}
%Recently, a number of datasets have been proposed for machine comprehension. Our task is related to that of \newcite{richardson2013mctest}, who have proposed \textit{MCTest}, a small curated dataset of 660 stories, with 4 multiple choice questions per story. The stories are crowdsourced and not limited to a domain. Answering questions in \textit{MCTest} requires drawing inferences from multiple sentences from the text passage. In our dataset, in contrast, answering some questions requires drawing inferences using knowledge not explicit in the text. \newcite{rajpurkar2016squad} have proposed the \textit{Stanford Question Answering Dataset} (\textit{SQuAD}), a data set of 100,000 questions on \textit{Wikipedia} articles collected via crowdsourcing. In SQuAD, the answer to a question corresponds to a segment/span from the reading passage. Since \textit{Wikipedia} articles mostly contain factual knowledge, \textit{SQuAD} does not assess how in practice, language comprehension relies on implicit and underrepresented knowledge about everyday activities i.e. script knowledge. \newcite{weston2015towards} have proposed the \textit{BAbI} dataset. \textit{BAbI} is a synthetic reading comprehension data set testing different types of reasoning to solve different tasks. In contrast to our dataset, the artificial texts in \textit{BAbI} are not reflective of a typically occurring narrative text.		

\section{Summary}
We present a new dataset for the task of machine comprehension focussing on commonsense knowledge. Questions were collected based on script scenarios, rather than  individual texts, which resulted in question--answer pairs that explicitly involve commonsense knowledge. In contrast to previous evaluation tasks, this setup allows us for the first time to assess the contribution of script knowledge for computational models of language understanding in a real-world evaluation scenario.

%We showed that the dataset covers a wide variety of question types. Our mode of collection results in a substantial number of questions that can only be answered via inference using commonsense knowledge. 
We expect our dataset to become a standard benchmark for testing models of commonsense and script knowledge. Human performance shows that the dataset is highly reliable. The results of several baselines, in contrast, illustrate that our task provides challenging test cases for the broader natural language processing community. MCScript forms the basis of a shared task organized at SemEval 2018. The dataset is available at \url{http://www.sfb1102.uni-saarland.de/?page_id=2582}.

\section*{Acknowledgements}
We thank the reviewers for their helpful comments. We also thank Florian Pusse for the help with the MTurk experiments and our student assistants Christine Schäfer, Damyana Gateva, Leonie Harter, Sarah Mameche, Stefan Grünewald and Tatiana Anikina for help with the annotations. This research was funded by the German Research Foundation (DFG) as part of SFB~1102 `Information Density and Linguistic Encoding' and EXC~284 `Multimodal Computing and Interaction'.

% \nocite{*}
\section{Bibliographical References}
\label{main:ref}

\bibliographystyle{lrec}
\bibliography{references}

\end{document}